\documentclass[
twocolumn,
]{ceurart}

\usepackage{inconsolata}
\usepackage{microtype}
\usepackage[frozencache]{minted}[json]
\usepackage{listings}
\ExpandArgs{c}\newcommand{new@fontshape}{}
\usepackage{gb4e}
\usepackage{tabularx}
\usepackage{booktabs}
\usepackage{multirow}
\usepackage{multicol}
\usepackage{relsize}
\newcolumntype{C}{>{\centering\arraybackslash}X}
\newcommand{\adhoc}{\textit{ad hoc}}
\newcommand{\etc}{\textit{etc.}}
\newcommand{\eg}{\textit{e.g.},}
\newcommand{\ie}{\textit{i.e.},}
\newcommand{\ia}{\textit{inter alia}}
\usepackage{enumitem}

\begin{document}

\copyrightyear{2024}
\copyrightclause{Copyright for this paper by its authors.
  Use permitted under Creative Commons License Attribution 4.0
  International (CC BY 4.0).}

\conference{CLiC-it 2024: Tenth Italian Conference on Computational Linguistics, Dec 04 — 06, 2024, Pisa, Italy}

\title{Nominal Class Assignment in Swahili}
\subtitle{A Computational Account}

\author[1]{Giada Palmieri}[%
email=giada.palmieri5@unibo.it,
url=https://giadapalmieri.github.io/,
]
\cormark[1]
\address[1]{University of Bologna}
\address[2]{Aalto University}
\author[2,1]{Konstantinos Kogkalidis}[%
email=kokos.kogkalidis@aalto.fi,
url=https://konstantinoskokos.github.io/,
]
\cormark[1]

\renewcommand{\arraystretch}{0.9}

\cortext[1]{Equal contribution. Authorship order was determined through a first-to-five game of rock paper scissors.}

\begin{abstract}
We discuss the open question of the relation between semantics and nominal class assignment in Swahili. We approach the problem from a computational perspective, aiming first to quantify the extent of this relation, and then to explicate its nature, taking extra care to suppress morphosyntactic confounds. Our results are the first of their kind, providing a quantitative evaluation of the semantic cohesion of each nominal class, as well as a nuanced taxonomic description of its semantic content.
\end{abstract}


\maketitle

\section{Introduction}
Swahili has a grand total of 18 nominal classes (\ie{} `genders').
There is no consensus on the extent to which the assignment of a noun to a given class is determined by its semantic content.
We explore this question from a computational angle.
Our experiments suggest semantic cohesion among nominal classes, and provide a summary of the taxonomic concepts associated to each class.

\section{Background}
\label{sec:background}

\subsection{Nominal Classes in Swahili}
Like other Bantu languages, Swahili has a rich nominal system, where nouns belong to different classes \cite{wald2018swahili, katamba2003bantu}, sometimes also referred to as `genders' \cite{spinner2014l2}.
The nominal class is signalled by an affix on the noun itself, and co-referenced with other elements of the sentence through grammatical agreement \cite{dixon1968noun}.

In Swahili, verbs require markers that agree with the nominal class of the subject.
An example of subject concord is reported below in (\ref{ex:1}): the noun \textit{mtoto} `child' bears the prefix of noun class 1 \textit{m-} on the noun, and agrees with the verb through the subject marker \textit{a}-. 
The same process can be observed in (\ref{ex:2}) for the noun \textit{mti} `tree' (class 3), or in (\ref{ex:3}) for \textit{kitabu} `book' (class 7).\footnote{Abbreviations used in the examples: [n] = nominal class; \textsc{sm} = subject marker; \textsc{prf} = perfect; \textsc{fv} = final vowel.}
\ea\label{ex:1} 
\gll M-toto a-me-anguk-a. \\
[1]-child \textsc{sm}[1]-\textsc{prf}-fall-\textsc{fv} \\ 
\glt `The child has fallen.'%
\ex\label{ex:2}
\gll M-ti  u-me-anguk-a. \\
[3]-tree  \textsc{sm}[3]-\textsc{prf}-fall-\textsc{fv} \\ 
\glt `The tree has fallen.'
\ex\label{ex:3}
\gll Ki-tabu  ki-me-anguk-a. \\
[7]-book  \textsc{sm}[7]-\textsc{prf}-fall-\textsc{fv} \\ 
\glt `The book has fallen.'
\z

Table \ref{tab:subj-concord} provides an overview of Swahili nominal classes, with their respective nominal affixes and subject concord markers.
The division of the nominal classes is based on reconstructions from Proto-Bantu \cite[\ia]{meeussen1967bantu, Guthrie1971comparative}, and it aims at maintaining a correspondence across Bantu languages.
Swahili is considered to have a total of 18 nominal classes, but some are missing in standard Swahili (\eg{} classes 12, 13 and 18), while others are not uniquely identified by their nominal affix and/or subject concord markers.
Odd numbers are traditionally associated with singular classes, and even numbers with plural classes.
The first ten classes are in singular/plural pairing relations (\eg{} class 2 is the plural form of class 1), while some singular noun classes may lack a plural form or borrow their plural forms from other classes. 

\begin{table}
  \caption{Swahili nominal classes.}
  \label{tab:subj-concord}
  \begin{tabular}{@{}ccc@{}}
    Nominal Class&Noun Affix&Subject Concord\\
    \toprule
    1/2 &  m-/wa-     &a-/wa-\\
    3/4 &  m-/mi-     &u-/i- \\
    5/6 &  (ji-)/ma-     &li-/ya-\\
    7/8 &   ki-/vi-    &ki-/vi- \\
    9/10 &  $\emptyset$    &i-/zi- \\
    11  &   u-    &u- \\
    14  &  u- &u- \\
    15  & ku-  &ku-\\
    16  & -ni  &pa- \\
    17  & -ni  &ku-\\
\end{tabular}
\end{table}

There is a long-standing debate on whether Bantu nominal classification is arbitrary~\cite{richards1967linguistic}, or whether it is based on some underlying semantic principles, with specific meanings associated to specific classes~\cite{zawawi1979loan,denny1986semantics}.
For Swahili, contemporary studies often adopt a stance that lies between these two extremes: nominal classification seems somewhat predictable based on semantic content, though it may often seem arbitrary~\cite{katamba2003bantu,krifka2005swahili,wald2018swahili,Marten2021classes}.
This view is also commonly found in textbooks: semantic cues are provided as an aid for the acquisition of Swahili, but accompanied by the admonition that many nouns do not necessarily admit generalizations~\citep{wilson1985simplified, safari2012swahili}.

\begin{figure}
\caption{\citeauthor{contini1994noun}'s semantic network for class 3.}
\label{fig:class3}
    \centering
    \includegraphics[width=1\linewidth]{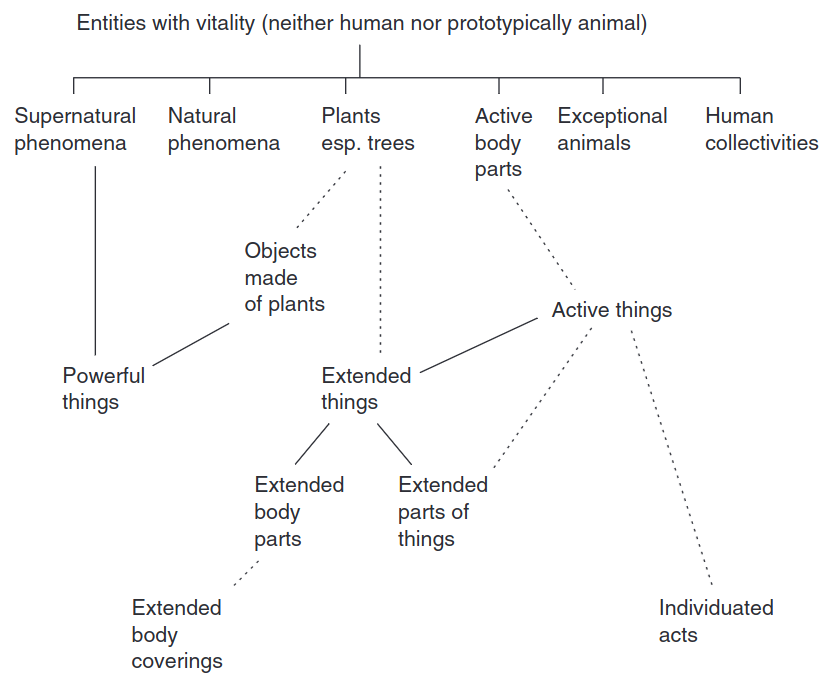}
\end{figure}

Two prominent attempts to examine the semantic categories associated with Swahili nominal classes are provided by \citet{contini1994noun} and \citet{moxley1998semantic}.
Both studies are cast in a cognitive linguistic framework, and propose networks of meanings and semantic features based on criteria such as resemblance or metaphoric and metonymic extensions. 
As an example, consider the semantic network for class 3 suggested by \citet{contini1994noun} in Figure \ref{fig:class3}: part of the branching includes the features
\textsc{plants}  $>$ \textsc{objects made of plants}  $>$ \textsc{powerful things}.
Similarly, \citet{moxley1998semantic} suggests a structure of class 3/4 where the notions of `plants, trees' extends to `parts of plants' or to objects with `long, thin, extended shape'.
These studies offer valuable insights into the principles underlying nominal classifications, suggesting the potential for more articulate generalizations than are immediately apparent.
However, note that they rely on features that were conceived \adhoc{} to account for the categorization of Swahili nouns.
Despite this, the nominal classification of several nouns remains unaccounted for \cite{katamba2003bantu}.
It is unclear whether this is due to features that were overlooked in these studies, or an indication that the classification of some nouns is inherently arbitrary.

\subsection{Computational Approaches to Swahili Nominal Classes}
Despite the long-standing theoretical debate, computational attempts at semantically characterizing Swahili nominal classes are few and far between.
In the context of word sense disambiguation, \citet{ng2005word} utilizes a collection of manually selected morphosyntactic features in combination with a self-organizing map in order to semantically cluster Swahili nouns.
The study finds that including noun prefix features (\ie{} nominal class indicators) moderately improves clustering performance, indicating a degree of coherence between semantics and morphology.
This improvement is particularly notable for classes 1/2, 7/8, and 11.
\citet{olstad2011noun} trains a naive Bayes classifier over a private, manually annotated dataset that specifically and explicitly marks the features proposed by \citet{contini1994noun}.
The approach is framed as an empirical test of \citeauthor{contini1994noun}'s hypothesis, which the trained model is claimed to experimentally confirm; nonetheless, this assessment is compromised by lukewarm results and a flawed evaluation.%
\footnote{The key metrics reported are dataset-wide accuracy and per-class area-under-the-curve. Both are over-optimistic: the first tends to favor class-imbalanced datasets, whereas the latter ignores precision and obfuscates the predictive conflict of the competing classifiers.}
More recently, \citet{byamugisha2022noun} builds a noun class disambiguation system for Runyankore, another Bantu language.
The system relies on both a morphological and a semantic component, the latter employing k-NN clustering of word vectors to resolve ambiguities that extend beyond nominal morphology. 
The work is results-oriented, adopting a task-driven NLP posturing -- its only tangible contribution is the system itself.

\section{Methodology}
Unlike prior works, we are neither interested in preemptively adopting or verifying some existing theory, nor in maximizing discriminative performance metrics in some artificial downstream task.
What we \textit{are} interested in is computationally investigating whether semantic content alone is indeed a predictor of nominal class membership.
At first glance, word vectors seem to make for a natural starting point.
However, language-native word vectors are bound to carry implicit morphological cues, trivializing the mapping to nominal classes (at worst), or obfuscating its semantic aspect (at best).
Word vectors (both distributional and predictive) are built on the basis of co-occurrence contexts and/or statistics.
The effect of grammatical agreement is that nouns will inadvertently co-occur with verbs that carry subject markers indicative of the noun's class.
Case in point, the examples in (\ref{ex:1}), (\ref{ex:2}) and (\ref{ex:3}) contain morphologically distinct entries of the same verbal stem, which disclose the subject's nominal class.
The same problem is expounded when using modern segmentation techniques which implicitly account for morphology by incorporating information at the sub-word (\ie{} syllable- or character-) level (\textit{cf.} \textsc{BPE}~\cite{gage1994new}, \textsc{SentencePiece}~\cite{kudo-richardson-2018-sentencepiece}, \ia{}).
To bypass the problem, we conduct our analyses on English translations of Swahili nouns.
Mediating meaning through a foreign language carries the risk of inducing translation shifts and introducing inaccuracies.
That said, we deem it a necessary compromise; the bottleneck completely erases any traces of morphology, which would otherwise confound our results (and their interpretation). 

\subsection{Data}
We first compile a list of nominal lexical entries by consulting the TUKI Swahili-English dictionary.
We gather these by scraping the dictionary's online version\footnote{Available at \url{https://swahili-dictionary.com}.}, filtering for pages under the category of Swahili nouns.
The scrape yields 5\,974 lexical entries.
Each lexical entry corresponds to a Swahili nominal homograph.
Each homograph is assigned one or more meanings, grouped under one or more subject concord classes.
Meanings are provided in English, in the form of (lists of) synonyms, brief descriptions, or mixtures of the two.
These are sometimes interlaced with linguistic metadata such as usage examples, apothegms, explanatory comments, \etc{}

\begin{figure}
\caption{Example of parsed lexical records.}
\label{fig:parsed-json}
\begin{minted}[
    fontsize=\footnotesize,
    escapeinside=||,
]{json}
[|...|
  {"entry": "yahe",
   "definition": "friend, comrade",
   "subject_concord": "a-/wa-"},
  {"entry": "yahe",
   "definition": "commoner",
   "subject_concord": "a-/wa-"},
|...|]
\end{minted}
\end{figure}

\begin{figure}
\caption{Occurrence counts of subject concord classes.}
\label{fig:concord-distribution}
    \centering
    \includegraphics[width=1\linewidth]{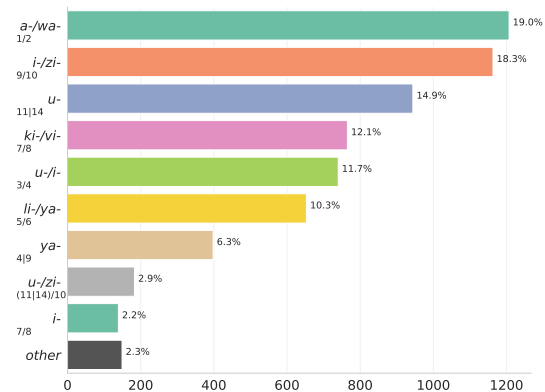}
\end{figure}

The dictionary is consistent in its typographic notation, which allows us to standardize its presentation with a tiny rule-based parser.
The parser removes metadata and splits homographs to nominals with unique meanings, gracefully pointing out the occasional inconsistency or error.
Guided by the parser, we identify and manually fix common typographic errors.
Following our corrections, we are left with a set of 6\,341 unique \textit{records}, \ie{} triplets of an entry identifier, a meaning and a subject concord class (Figure~\ref{fig:parsed-json}).
The distribution of subject concord classes is heavily skewed (Figure~\ref{fig:concord-distribution}).
We keep records assigned to one of the 9 most populous classes, which together account for about 98\% of the data, and discard the rest.
In what follows, we use these subject concord markers as an approximation of the underlying nominal classes.%
\footnote{The use of subject concord markers over noun affixes is mandated by the annotation format of the TUKI dictionary.}
The records we are left with correspond to the nominal classes 1/2, 3/4, 5/6, 7/8, 9/10, 11|14, 4|9 and (11|14)/10; the latter three are necessarily conflated or ambiguous due to their shared morphology.%
\footnote{We use the pipe operator ($\cdot$|$\cdot$) to denote disjunction.}

\subsection{Predicting Nominal Classes with a Language Model}
Our data allows for a first quantitative inquiry into the semantic uniformity and separation of nominal classes.
For our first take, we employ a supervised learning approach.
We task a small language model with predicting a record's subject concord class through the phrasal representation of its English definition.
The use of a pretrained language model allows the seamless representation of translations that are not strict word-to-word correspondences, promising also the ability to capture subtle semantic distinctions in the process.

\begin{table*}[h]
\caption{Macro- and micro-averaged and per-class F1 scores.}
\label{tab:f1}
    \newcommand{\num}[2]{\ensuremath{#1}\smaller\ensuremath{\textcolor{gray!90}{\pm{#2}}}}
    \centering
    \smaller
    \begin{tabularx}{\linewidth}{@{}CCCCCCCCCCC@{}}
        \textsc{M} &
            $\mu$ &
            \textit{a-/wa-} & 
            \textit{i-/zi-} & 
            \textit{u-} & 
            \textit{ki-/vi-} & 
            \textit{u-/i-} & 
            \textit{li-/ya-} & 
            \textit{ya-} &
            \textit{u-/zi-} & 
            \textit{i-}
        \\
         \toprule
         \num{34.5}{2.6} & 
            \num{48.6}{1.4} &
            \num{89.4}{0.5} & 
            \num{35.3}{2.9} & 
            \num{60.0}{3.9} & 
            \num{30.2}{2.2} & 
            \num{42.2}{6.2} & 
            \num{24.5}{4.2} &
            \num{21.4}{10.2} &
            \num{\hphantom{0}5.8}{11.4} &
            \num{\hphantom{0}1.8}{5.1}        
    \end{tabularx}
\end{table*}

\begin{table}
\caption{Confusion matrix over subject concord predictions.}
\label{tab:cm}
    \centering
    \includegraphics[width=1\linewidth]{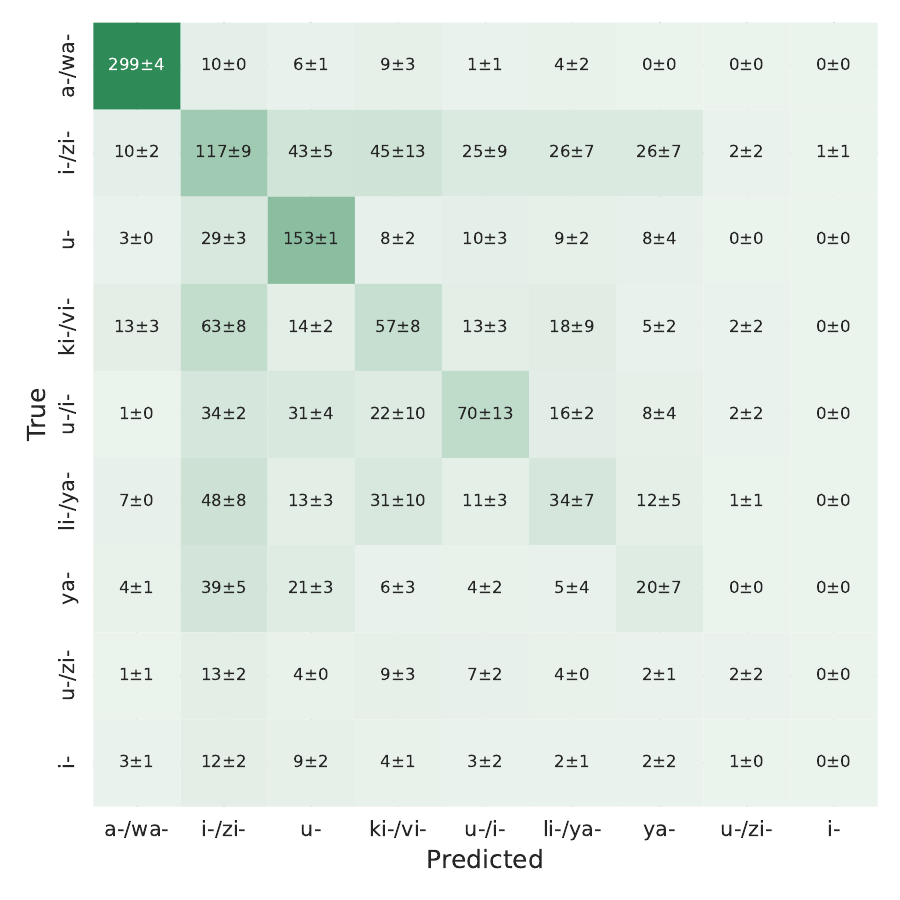}
\end{table}

We use \textsc{MiniLM}v2~\cite{wang2021minilmv2}, a distilled encoder-only model that has been fine-tuned for sentential similarity using a contrastive learning objective.
We apply a 75/25 train/eval split and further fine-tune the model to the task (we follow standard practices, attaching a neural classifier to the model's topmost layer, applied exclusively on the start-of-sequence token).
Model selection is based on evaluation loss; we select three models from as many training repetitions over the same split (one model per repetition).

We report means and 95\% confidence intervals for the macro- and micro-averaged and per-class F1 scores in Table~\ref{tab:f1}, and per-class predictions in Table~\ref{tab:cm}.
Across repetitions, the model is quick to fit the training set, but struggles to generalize, especially on under-represented classes.
Despite the fact, performance is significantly better than a probability-weighted random baseline (macro F1 of 14.3).

\subsection{Finding the Taxonomies of Nominal Classes with WordNet}
Our mixed results paint a nuanced picture.
Performance above random affirms that nominal classes are to an extent semantically coherent -- even if not \textit{perfectly} so.
Performance below perfect, however, offers nothing tangible.
The model's shortcomings might be indicative of a semantic dispersion or arbitrariness within nominal classes, but could also be attributed to the model itself, the training process, or the dataset.
In either case, we have strong evidence of an (at least partial) overlap between (at least some) semantic and morphological clusters.
Other than this confirmation, the supervised approach does not have much else to offer at this stage; over-parameterized black-box models are notoriously hard to extract linguistic insights from.
To actually \textit{ascribe} semantic descriptions to nominal classes, we need a better behaved alternative.
For our second take, we employ an unsupervised topic modeling approach.
We turn to WordNet~\cite{miller1995wordnet}, a lexical database that maps words to \textit{synsets}: semantically equivalent senses, equipped with periphrastic definitions that are linked together by binary semantic relations.

We begin by matching Swahili records with English WordNet synsets%
\footnote{A `native' WordNet would be a better fit for the task, but no mature Swahili version exists as of the time of writing.}.
Matching on a lexical basis is once again impossible; there is no natural correspondence between Swahili nouns and English synsets.
As a workaround, we use the same off-the-shelf language model (this time without any additional fine-tuning) to procure semantic representations of Swahili records and English synsets using their respective definitions.
We compute a matrix of pairwise scores in the Cartesian product of records and synsets with cosine similarity as our metric.
For each Swahili record we then isolate the most similar synsets -- no more than 10, and with a similarity score of no less than 0.5.
These enact entry points for the Swahili record into the WordNet graph.
For each synset, we extract all its \textit{hypernymy paths}: synset sequences that correspond to progressively broader taxonomic generalizations.
The \textit{meet} of hypernymy paths originating from multiple synsets associated to a single record correspond to all possible hypernyms of that record.
For each record, we weight hypernyms according to their occurrence counts divided by the total number of hypernymy paths in the record; intuitively, hypernyms are assigned a higher weight the more paths pass through them.
The process is noisy: error sources include both the matching, and WordNet itself.
Nonetheless, we are less interested in the hypernyms of individual records, and more so in their distribution across nominal classes.

\begin{table*}[h]
\caption{Macro-averaged and per-class weighted relevance between taxonomic descriptors and nominal classes.}
\label{tab:relevance}
    \newcommand{\num}[2]{\ensuremath{#1}\smaller\ensuremath{\textcolor{gray!90}{\pm{#2}}}}
    \centering
    \smaller
    \begin{tabularx}{.85\linewidth}{@{}CCCCCCCCC@{}}
            \textit{a-/wa-} & 
            \textit{i-/zi-} & 
            \textit{u-} & 
            \textit{ki-/vi-} & 
            \textit{u-/i-} & 
            \textit{li-/ya-} & 
            \textit{ya-} &
            \textit{u-/zi-} & 
            \textit{i-}
        \\
         \toprule
            0.102 &
            0.018 &
            0.040 &
            0.017 &
            0.025 & 
            0.016 &
            0.016 &
            0.014 &
            0.009     
    \end{tabularx}
\end{table*}

On the basis of the above, we have access to the joint probability of nominal classes and hypernyms, $p_{\textsc{c}\times\textsc{h}}$, as well as their marginal probabilities, $p_{\textsc{c}}$ and $p_{\textsc{h}}$.
We filter out hypernyms with less than 10 global occurrences, and compute the frequency-weighted%
\footnote{The scaling helps alleviate the `rare event' bias of vanilla PMI.} pointwise mutual information between classes and hypernyms:
\begin{equation}
    \mathrm{wPMI}(c, h) := p_{\textsc{c}\times\textsc{h}}(c,h)~\mathrm{PMI}(c, h)
\end{equation}
where:
\begin{equation}
     \mathrm{PMI}(c,h) := log_2\left(\frac{p_{\textsc{c}\times\textsc{h}}(c, h)}{p_{\textsc{c}}(c)p_{\textsc{h}}(h)}\right)
\end{equation}
Pairs with a positive wPMI score indicate \textit{relevance} (\ie{} mutual dependence) between their coordinates -- the higher the score, the better a hypernym \textit{describes} a subject concord class.
The aggregation of positive scores allows us to quantify and compare the semantic cohesion of subject concord classes given their descriptions -- we present these in Table~\ref{tab:relevance}.
We also present the top 20 extracted descriptors along with their scores in Appendix~\ref{appendix-table}.
The sum total of positive mutual information between extracted descriptors and subject concord classes under this weighting scheme is approximately 0.26 shannons, suggesting a moderate bidirectional dependency between the two.

\section{Analysis}
For several classes, our experimental results are congruent with the hypotheses of \citet{contini1994noun} and \citet{moxley1998semantic}, \ia{}. Concretely:
\begin{itemize}[topsep=0pt,leftmargin=*,noitemsep]
    \item Subject concord class \textit{a-/wa-} is associated with \textbf{humans}, \textbf{causal agents} and \textbf{animacy}; the class is the most semantically coherent and categorically defined; the classifier can accurately predict it, and its taxonomic descriptors are well-pronounced.
    \item Subject concord class \textit{u-} predominantly refers to \textbf{abstract concepts}; the class is the second easiest to predict, and has the most homogeneous description.
    \item Subject concord class \textit{u-/i-} is mostly associated with \textbf{plants}; it is the third easiest class to predict, but predictions are already getting somewhat unreliable.
    \item Subject concord class \textit{i-/zi-} is semantically \textbf{disparate}; its descriptors are heterogeneous and carry relatively low scores. This disparity is consistent with the class' characterization as a `residual catchall category'~\cite{zawawi1979loan,contini1994noun} where loanwords are often assigned~\cite{schadeberg2009loanwords}. The only standout descriptor relates the class to \textbf{human-made objects}, but the same descriptor dominates also classes \textit{li-/ya-} and \textit{ki-/vi-}.%
    \footnote{Describing \textit{li-/ya} and \textit{ki-/vi-} as human-made objects is in partial alignment with the literature. The two are respectively associated with `augmentative' and `dimininutive' meanings~\cite{moxley1998semantic} and, by extension, with big or small objects~\cite{contini1994noun}.}
    Indeed, the model struggles to tell these three classes apart.
\end{itemize}
\vspace{0.2em}

In addition to experimentally affirming existing hypotheses, our approach also yields novel insights and artifacts.
With respect to \textit{ya-} and \textit{i-}, the macro-level summary of these two understudied classes reveals an as-of-yet undocumented pattern:
both classes lack a singular-plural paradigm, and contain concepts broadly categorized as \textbf{abstractions}, albeit of different kinds.
This observation may support the correlation between uncountability and abstract meanings noticed in other languages~\cite{Katz2012QuantifyingCE, husic2020abstract}; doing so would however require a thorough examination of these nouns' properties.

From a high-level perspective, we have chosen to isolate the first few highest-ranked semantic components of each class.
This ensures backwards compatibility with the literature, but is also a very radical simplification. 
In reality, our descriptions are fine-grained enough to allow semantically distinguishing between any two classes, even when their primary descriptors overlap.
Case in point, \textit{i-/zi-}, \textit{ki-/vi-} and \textit{li-/ya-} have all been reduced to `human-made objects'; yet the three are actually very different, having only 2 (out of a total of 41) descriptors in common.
Moreover, a descriptor is not just a (weighted) concept in isolation, but inherits also the expansive structure of the underlying WordNet it came from.
In that sense, our approach does not only describe nominal classes with WordNet synsets, but dually also decorates the WordNet graph with nominal class weights.

\section{Conclusions}
We explored the relation between semantics and nominal class assignment in Swahili.
We approached the question from two complementary computational angles.
Verifying first the presence of a relation using supervised learning, we then sought to explicate its nature using unsupervised topic modeling.
Starting from a blank slate and without any prior interpretative bias, our methodology rediscovered go-to theories of Swahili nominal classification, while also offering room for further insights and explorations.
Our work is among the first to tackle Bantu nominal assignment computationally, and the first to focus exclusively on semantics.
Our methodology is typologically unbiased and computationally accessible, allowing for an easy extension to other languages, under the sole requirement of a dictionary.
We make our scripts and generated artifacts publicly available at \url{https://github.com/konstantinosKokos/swa-nc}.

We leave several directions open to future work.
We have experimented with a single dataset, a single model and a single lexical database; varying either of these coordinates and aggregating the results should  help debias our findings.
We have only looked for semantic generalizations across hyperonymic taxonomies -- looking at other kinds of lexical relations might yield different semantic observations.
Our chosen metric of relevance is by construction limited to first-order pairwise interactions, failing to account for exceptional cases or conditional associations.
Finally, we had to resort to computational acrobatics through English in order to access necessary tools and resources.
This is yet another reminder of the disparities in the pace of `progress' of language technology, and a call for the computational inclusion of typologically diverse languages.

\section{Acknowledgments}
We are grateful to Joost Zwarts and to three anonymous reviewers for their helpful feedback.

\bibliography{swahili}

\appendix
\twocolumn[
\section{Appendix}\label{appendix-table}
Taxonomic description of nominal classes. 
Scores are multiplied by 100$p_{\textsc{c}}(c)^{-1}$ to enhance legibility and facilitate direct numerical comparison across classes.
Bold face scores indicate higher mutual information. Grayed out descriptors are hyponyms of at least one other descriptor with a higher score.\\

\begin{tabularx}{0.99\linewidth}{@{}lX@{}}
    Subject Concord & Top 20 Descriptors\\
    \toprule
    \textit{a-/wa-} & person.n.01~(\textbf{8.5}), organism.n.01~(\textbf{5.8}), living\_thing.n.01~(\textbf{5.8}), causal\_agent.n.01~(\textbf{4.1}), physical\_entity.n.01~(\textbf{3.3}), \textcolor{gray!80}{animal.n.01~(\textbf{2.9})}, \textcolor{gray!80}{chordate.n.01~(\textbf{2.3})}, \textcolor{gray!80}{vertebrate.n.01~(\textbf{2.3})}, whole.n.02~(\textbf{2.1}), \textcolor{gray!80}{object.n.01~(\textbf{1.6})}, \textcolor{gray!80}{bird.n.01~(0.8)}, \textcolor{gray!80}{aquatic\_vertebrate.n.01~(0.7)}, \textcolor{gray!80}{fish.n.01~(0.7)}, taxonomic\_group.n.01~(0.7), biological\_group.n.01~(0.7), \textcolor{gray!80}{adult.n.01~(0.6)}, \textcolor{gray!80}{bad\_person.n.01~(0.6)}, \textcolor{gray!80}{mammal.n.01~(0.5)}, \textcolor{gray!80}{unwelcome\_person.n.01~(0.5)}, \textcolor{gray!80}{relative.n.01~(0.5)}\\
    \addlinespace
    \textit{i-/zi-} & artifact.n.01~(\textbf{1.2}), abstraction.n.06~(0.6), \textcolor{gray!80}{instrumentality.n.03~(0.6)}, matter.n.03~(0.3), \textcolor{gray!80}{device.n.01~(0.3)}, \textcolor{gray!80}{measure.n.02~(0.3)}, \textcolor{gray!80}{communication.n.02~(0.3)}, \textcolor{gray!80}{substance.n.07~(0.2)}, \textcolor{gray!80}{food.n.01~(0.2)}, \textcolor{gray!80}{relation.n.01~(0.2)}, \textcolor{gray!80}{implement.n.01~(0.2)}, clothing.n.01~(0.2), \textcolor{gray!80}{fundamental\_quantity.n.01~(0.1)}, \textcolor{gray!80}{time\_period.n.01~(0.1)}, color.n.01~(0.1), \textcolor{gray!80}{possession.n.02~(0.1)}, entity.n.01~(0.1), \textcolor{gray!80}{chromatic\_color.n.01~(0.1)}, \textcolor{gray!80}{substance.n.01~(0.1)}, visual\_property.n.01~(0.1)\\
    \addlinespace
    \textit{u-} & abstraction.n.06~(\textbf{5.5}), \textcolor{gray!80}{attribute.n.02~(\textbf{3.9})}, \textcolor{gray!80}{psychological\_feature.n.01~(\textbf{2.3})}, \textcolor{gray!80}{event.n.01~(\textbf{1.7})}, \textcolor{gray!80}{act.n.02~(\textbf{1.5})}, \textcolor{gray!80}{state.n.02~(\textbf{1.4})}, \textcolor{gray!80}{quality.n.01~(\textbf{1.4})}, entity.n.01~(\textbf{1.2}), \textcolor{gray!80}{trait.n.01~(0.7)}, \textcolor{gray!80}{activity.n.01~(0.7)}, \textcolor{gray!80}{cognition.n.01~(0.6)}, \textcolor{gray!80}{property.n.02~(0.6)}, \textcolor{gray!80}{feeling.n.01~(0.5)}, \textcolor{gray!80}{condition.n.01~(0.5)}, \textcolor{gray!80}{group\_action.n.01~(0.4)}, \textcolor{gray!80}{action.n.01~(0.4)}, \textcolor{gray!80}{change.n.03~(0.3)}, \textcolor{gray!80}{process.n.02~(0.2)}, \textcolor{gray!80}{work.n.01~(0.2)}, \textcolor{gray!80}{immorality.n.01~(0.2)}\\
    \addlinespace
    \textit{ki-/vi-} & artifact.n.01~(\textbf{2.1}), \textcolor{gray!80}{instrumentality.n.03~(\textbf{1.1})}, object.n.01~(1.0), physical\_entity.n.01~(0.9), \textcolor{gray!80}{whole.n.02~(0.7)}, \textcolor{gray!80}{device.n.01~(0.6)}, part.n.03~(0.5), \textcolor{gray!80}{thing.n.12~(0.5)}, \textcolor{gray!80}{body\_part.n.01~(0.5)}, \textcolor{gray!80}{structure.n.01~(0.3)}, symptom.n.01~(0.2), evidence.n.01~(0.2), \textcolor{gray!80}{container.n.01~(0.2)}, \textcolor{gray!80}{covering.n.02~(0.2)}, information.n.02~(0.2), \textcolor{gray!80}{implement.n.01~(0.2)}, communication.n.02~(0.2), \textcolor{gray!80}{clothing.n.01~(0.2)}, relation.n.01~(0.2), \textcolor{gray!80}{location.n.01~(0.2)}\\
    \addlinespace
    \textit{u-/i-} & plant.n.02~(\textbf{2.6}), \textcolor{gray!80}{vascular\_plant.n.01~(\textbf{2.6})}, \textcolor{gray!80}{woody\_plant.n.01~(\textbf{2.0})}, \textcolor{gray!80}{tree.n.01~(\textbf{1.6})}, event.n.01~(0.7), \textcolor{gray!80}{happening.n.01~(0.5)}, whole.n.02~(0.5), dicot\_genus.n.01~(0.5), object.n.01~(0.5), \textcolor{gray!80}{angiospermous\_tree.n.01~(0.4)}, psychological\_feature.n.01~(0.4), wood.n.01~(0.4), plant\_material.n.01~(0.4), \textcolor{gray!80}{herb.n.01~(0.4)}, \textcolor{gray!80}{shrub.n.01~(0.3)}, \textcolor{gray!80}{sound.n.04~(0.3)}, action.n.01~(0.3), \textcolor{gray!80}{change.n.03~(0.3)}, material.n.01~(0.3), \textcolor{gray!80}{act.n.02~(0.3)}\\
    \addlinespace
    \textit{li-/ya-} & artifact.n.01~(\textbf{1.5}), object.n.01~(0.9), physical\_entity.n.01~(0.8), \textcolor{gray!80}{instrumentality.n.03~(0.7)}, \textcolor{gray!80}{whole.n.02~(0.5)}, \textcolor{gray!80}{thing.n.12~(0.5)}, \textcolor{gray!80}{part.n.03~(0.4)}, \textcolor{gray!80}{matter.n.03~(0.4)}, \textcolor{gray!80}{body\_part.n.01~(0.4)}, \textcolor{gray!80}{structure.n.01~(0.4)}, \textcolor{gray!80}{natural\_object.n.01~(0.3)}, \textcolor{gray!80}{container.n.01~(0.3)}, edible\_fruit.n.01~(0.2), \textcolor{gray!80}{solid.n.01~(0.2)}, \textcolor{gray!80}{food.n.02~(0.2)}, plant\_organ.n.01~(0.2), \textcolor{gray!80}{plant\_part.n.01~(0.2)}, \textcolor{gray!80}{reproductive\_structure.n.01~(0.2)}, shape.n.02~(0.2), \textcolor{gray!80}{substance.n.01~(0.2)}\\
    \addlinespace
    \textit{ya-} & abstraction.n.06~(\textbf{3.3}), \textcolor{gray!80}{psychological\_feature.n.01~(\textbf{2.0})}, \textcolor{gray!80}{event.n.01~(\textbf{1.6})}, \textcolor{gray!80}{act.n.02~(\textbf{1.3})}, entity.n.01~(0.9), \textcolor{gray!80}{attribute.n.02~(0.7)}, \textcolor{gray!80}{speech\_act.n.01~(0.7)}, matter.n.03~(0.6), \textcolor{gray!80}{state.n.02~(0.6)}, \textcolor{gray!80}{relation.n.01~(0.5)}, \textcolor{gray!80}{group\_action.n.01~(0.5)}, \textcolor{gray!80}{communication.n.02~(0.4)}, \textcolor{gray!80}{cognition.n.01~(0.4)}, \textcolor{gray!80}{substance.n.01~(0.3)}, phenomenon.n.01~(0.3), process.n.06~(0.3), \textcolor{gray!80}{natural\_phenomenon.n.01~(0.3)}, \textcolor{gray!80}{activity.n.01~(0.3)}, \textcolor{gray!80}{feeling.n.01~(0.3)}, \textcolor{gray!80}{request.n.02~(0.3)}\\
    \addlinespace
    \textit{u-/zi-} & artifact.n.01~(\textbf{3.3}), object.n.01~(\textbf{2.9}), physical\_entity.n.01~(\textbf{2.4}), \textcolor{gray!80}{whole.n.02~(\textbf{1.9})}, \textcolor{gray!80}{thing.n.12~(\textbf{1.4})}, \textcolor{gray!80}{part.n.03~(\textbf{1.4})}, \textcolor{gray!80}{body\_part.n.01~(\textbf{1.2})}, \textcolor{gray!80}{instrumentality.n.03~(\textbf{1.2})}, \textcolor{gray!80}{implement.n.01~(0.7)}, palm.n.03~(0.6), \textcolor{gray!80}{part.n.02~(0.6)}, \textcolor{gray!80}{location.n.01~(0.5)}, \textcolor{gray!80}{natural\_object.n.01~(0.5)}, \textcolor{gray!80}{device.n.01~(0.5)}, body\_covering.n.01~(0.5), indefinite\_quantity.n.01~(0.5), \textcolor{gray!80}{hair.n.01~(0.5)}, \textcolor{gray!80}{decoration.n.01~(0.4)}, poem.n.01~(0.4), \textcolor{gray!80}{appendage.n.03~(0.4)}\\
    \addlinespace
    \textit{i-} & abstraction.n.06~(\textbf{3.2}), region.n.03~(\textbf{1.0}), location.n.01~(1.0), \textcolor{gray!80}{psychological\_feature.n.01~(0.9)}, matter.n.03~(0.7), \textcolor{gray!80}{cognition.n.01~(0.6)}, \textcolor{gray!80}{attribute.n.02~(0.6)}, entity.n.01~(0.6), \textcolor{gray!80}{substance.n.01~(0.6)}, \textcolor{gray!80}{district.n.01~(0.5)}, \textcolor{gray!80}{substance.n.07~(0.5)}, \textcolor{gray!80}{administrative\_district.n.01~(0.5)}, gathering.n.01~(0.5), \textcolor{gray!80}{relation.n.01~(0.5)}, \textcolor{gray!80}{state.n.02~(0.5)}, \textcolor{gray!80}{geographical\_area.n.01~(0.5)}, \textcolor{gray!80}{group.n.01~(0.5)}, \textcolor{gray!80}{condition.n.01~(0.5)}, process.n.06~(0.5), physical\_phenomenon.n.01~(0.5)
\end{tabularx}
]

\end{document}